\documentclass[10pt,twocolumn,letterpaper]{article}

\usepackage{wacv}              

\usepackage{graphicx}
\usepackage{amsmath}
\usepackage{amssymb}
\usepackage{booktabs}
\usepackage[table,xcdraw]{xcolor} 
\usepackage{booktabs} 
\usepackage{svg}

\newcommand{\bB}{\mathbf{B}}
\newcommand{\bc}{\mathbf{c}}\newcommand{\bC}{\mathbf{C}}

\newcommand{\bE}{\mathbf{E}}
\newcommand{\bG}{\mathbf{G}}

\newcommand{\bI}{\mathbf{I}}

\newcommand{\bo}{\mathbf{o}}

\newcommand{\bq}{\mathbf{q}}
\newcommand{\bR}{\mathbf{R}}
\newcommand{\bS}{\mathbf{S}}

\newcommand{\bx}{\mathbf{x}}

\newcommand{\bmu}{\boldsymbol{\mu}}

\newcommand{\bOmega}{\boldsymbol{\Omega}}

\newcommand{\nR}{\mathbb{R}}

\newcommand{\cL}{\mathcal{L}}

\newcommand{\figref}[1]{Fig.~\ref{#1}}

\newcommand{\tabnref}[1]{Table~\ref{#1}}

\makeatletter
\DeclareRobustCommand\onedot{\futurelet\@let@token\@onedot}
\def\@onedot{\ifx\@let@token.\else.\null\fi\xspace}
 
\def\ie{i.e\onedot}

\makeatother

\newcommand{\PAR}[1]{\vspace{0.1cm}\noindent{\bf #1} }

\newcommand{\norm}[1]{\left\lVert#1\right\rVert}

\usepackage[pagebackref,breaklinks,colorlinks]{hyperref}

\usepackage[capitalize]{cleveref}
\crefname{section}{Sec.}{Secs.}
\Crefname{section}{Section}{Sections}
\Crefname{table}{Table}{Tables}
\crefname{table}{Tab.}{Tabs.}

\def\wacvPaperID{*****} 
\def\confName{WACV}
\def\confYear{2025}

\begin{document}

\title{BeSplat: Gaussian Splatting from a Single Blurry Image and Event Stream\vspace{-1em}}

\author{Gopi Raju Matta\\
IIT Madras\\
{\tt\small ee17d021@smail.iitm.ac.in}
\and
Reddypalli Trisha\\
IIIT RGUKT RK Valley\\
{\tt\small trishareddypalli@gmail.com}
\and
Kaushik Mitra\\
IIT Madras\\
{\tt\small kmitra@ee.iitm.ac.in}
}

\twocolumn[{%
\maketitle
\vspace{-3em} 
\begin{center}
    \url{https://gopirajumatta.github.io/BeSplat/}
\end{center}
\vspace{0em} 
}]

\begin{figure*}[th]
    \centering
    \includegraphics[width=0.98\linewidth]{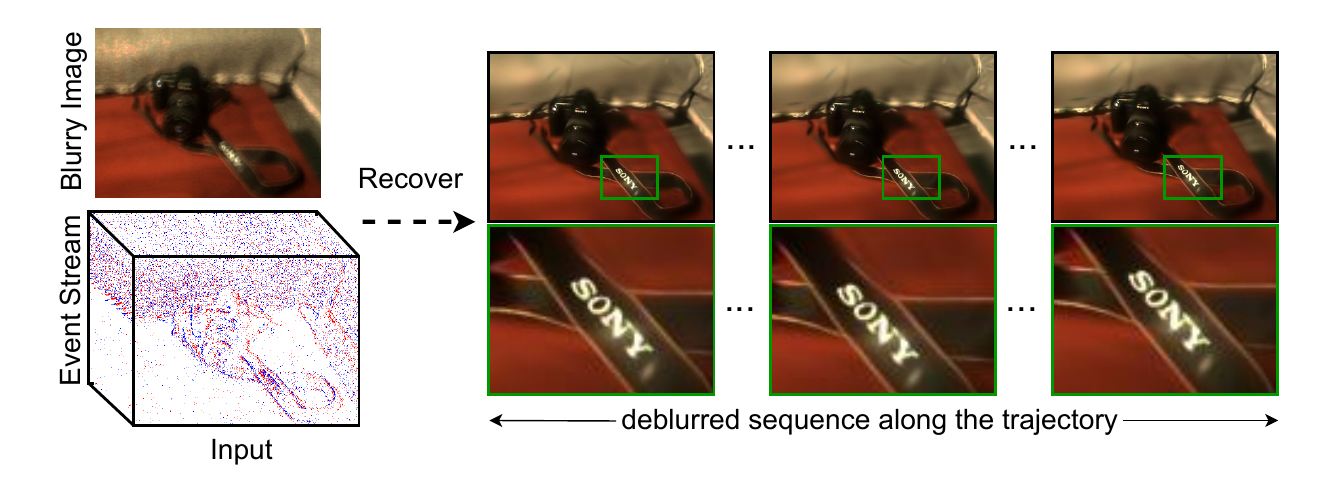}
    \caption{Given a single blurry image and its corresponding event stream, BeSplat synthesizes high-quality, novel sharp images along the camera trajectory by harnessing event information to precisely estimate motion trajectory and restore fine details with clarity.
    }
    \label{fig:method-overall}
\end{figure*}

\begin{abstract}
   Novel view synthesis has been greatly enhanced by the development of radiance field methods. The introduction of 3D Gaussian Splatting (3DGS) has effectively addressed key challenges, such as long training times and slow rendering speeds, typically associated with Neural Radiance Fields (NeRF), while maintaining high-quality reconstructions. 
   In this work (BeSplat), we demonstrate the recovery of sharp radiance field (Gaussian splats) from a single motion-blurred image and its corresponding event stream. 
   Our method jointly learns the scene representation via Gaussian Splatting and recovers the camera motion through Bézier SE(3) formulation effectively, minimizing discrepancies between synthesized and real-world measurements of both blurry image and corresponding event stream. We evaluate our approach on both synthetic and real datasets, showcasing its ability to render view-consistent, sharp images from the learned radiance field and the estimated camera trajectory. 
   To the best of our knowledge, ours is the first work to address this highly challenging ill-posed problem in a Gaussian Splatting framework with the effective incorporation of temporal information captured using the event stream.
\end{abstract}

\section{Introduction}
\label{sec:intro}

Acquiring accurate 3D scene representations from 2D images has long been a challenging problem in computer vision. This task is fundamental for various applications such as virtual and augmented reality, as well as robotics navigation. Substantial efforts have been dedicated to solving it over the years. Among the most notable advancements is the Neural Radiance Fields (NeRF) \cite{nerf}, which uses Multi-Layer Perceptrons (MLPs) and a differentiable volume rendering technique \cite{levoy1990efficient, max1995optical}. NeRF has garnered significant attention due to its ability to recover high-quality 3D scene representations from 2D images.

Numerous works have focused on improving NeRF's performance, particularly in terms of training \cite{mueller2022instant, Chen2022ECCV, yu_and_fridovichkeil2021plenoxels} and rendering efficiency \cite{Garbin2021, Yu2021}. A more recent approach, 3D Gaussian Splatting (3D-GS) \cite{kerbl3Dgaussians}, extends implicit neural rendering \cite{nerf} by representing scenes as explicit point clouds of Gaussians. By projecting these Gaussians onto the image plane, 3D-GS achieves real-time rendering, improving both training and rendering efficiency while also enhancing scene quality.

To address the deblurring challenge, several approaches have been proposed, including methods based on implicit neural representations like NeRF \cite{deblur-nerf, Lee_2023_CVPR, wang2023badnerf}. For instance, Deblur-NeRF \cite{deblur-nerf} introduces a deformable sparse kernel to simulate the blur, while DP-NeRF \cite{Lee_2023_CVPR} incorporates physical priors into Deblur-NeRF to recover a clean NeRF representation. BAD-NeRF \cite{wang2023badnerf} models the motion blur process and jointly optimizes NeRF while recovering the camera trajectory during exposure. However, these implicit methods struggle with real-time rendering and generating high-quality outputs, due to the complexity of their neural models. 

Event stream can be acquired by an event camera \cite{lichtsteiner2008128} which captures pixel intensity changes caused by the relative motion between the scene and camera. Unlike standard frame-based cameras, event camera captures asynchronous events with very low latency, leading to extremely high temporal resolution\cite{gallego2022eventbased}. This characteristic compensates with the image formation process of a blurry image (i.e. integral of photon measurements across time). Several prior works thus take advantage of both modalities for high quality single image deblurring \cite{pan2019bringing, wang2020event, sun2022eventbaseda}. However, these methods are unable to recover the camera motion trajectory and extract structural details from a single blurry image, thereby limiting their applicability in 3D computer vision tasks.

Some NeRF-based methods that incorporate event stream \cite{qi2023e2nerf, low2023_robust-e-nerf, klenk2022nerf, rudnev2023eventnerf, hwang2023ev} demonstrate the capability to achieve image deblurring and accurate reconstruction of neural radiance fields. Nonetheless, these methods necessitate input images from multiple viewpoints alongside event data.  

In contrast, BeNeRF \cite{li2025benerf} explores the usage of only a single blurry image along with event data to recover the sharp radiance field and the unknown camera motion. It represents the continuous camera motion with a cubic B-Spline in SE(3) space and define it as the trajectory of both frame-based camera and event camera. Given the neural 3D representation and interpolated poses from the cubic B-Spline, it can synthesize both the blurry image and the brightness change within a time interval via the physical image formation process. The NeRF and motion trajectory can then be jointly optimized by minimizing the difference between the synthesized data and the real measurements. However, being a NeRF based method, both the training and rendering times are large.

In this work, we propose a novel approach for motion deblurring within the Gaussian Splatting framework, which we call as \textbf{BeSplat}. Our method addresses the challenges associated with motion-blurred images by leveraging the advantages of Gaussian Splatting and the finer temporal resolution of event stream. We effectively integrate the temporal information captured by the event stream into the Gaussian Splatting framework, significantly improving the deblurring process. Unlike previous approaches, which require multi-view images and complex pose estimation, BeSplat uses a single blurry image and the corresponding event stream to perform deblurring and recover sharp 3D representations. We model the camera motion during the exposure time using a Bézier curve in SE(3) space, with 6-DoF poses interpolated from the Bézier curve of order 7. By utilizing Gaussian Splatting, we can render sharp views of the scene and reconstruct the camera motion trajectory that generated the blurry image.

While BeNeRF \cite{li2025benerf} performs a similar task using NeRF, our work applies this concept to Gaussian Splatting, benefiting from significantly accelerated training times and real-time rendering, while still producing comparable results. Gaussian Splatting's explicit point cloud representation enables faster and more efficient scene rendering compared to the implicit representations used in NeRF.

We conduct experiments on both synthetic and real datasets, demonstrating that BeSplat outperforms existing state-of-the-art methods for both motion deblurring and novel view synthesis, achieving real-time rendering with comparable quality.

To summarize, our main contributions are as follows:
\begin{itemize}
    \item Our work is the first to conceptualize single-image motion deblurring as a novel view synthesis problem within the Gaussian Splatting framework, effectively utilizing the complementary information offered by an event stream.
    \item We incorporated an event loss into the framework inspired from BeNeRF to address the ill-posed nature of single-image deblurring, enabling accurate estimation of the camera motion trajectory.
    \item Our method achieves accelerated training times, reduced GPU memory usage, and real-time rendering capabilities, while producing sharp, high-quality novel images along the camera trajectory. 
\end{itemize}

\begin{figure*}[th]
    \centering
    \includegraphics[width=0.98\linewidth]{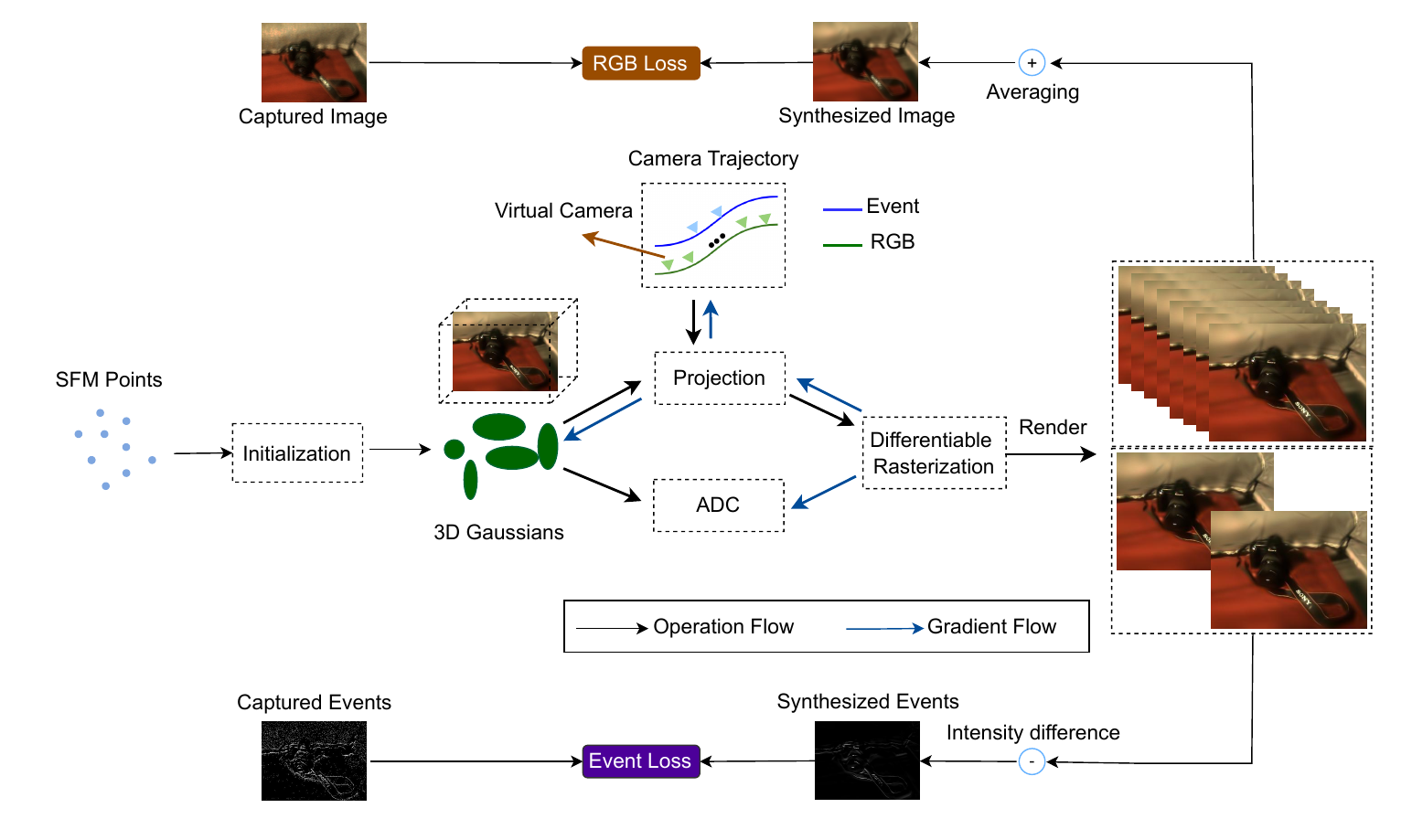}
  \caption{\textbf{Our Method:} The BeSplat framework reconstructs sharp radiance fields (Gaussian splats) while accurately estimating the camera motion trajectory, modeled with a Bézier curve, from a single blurry image and its corresponding event stream. The framework jointly optimizes the Gaussian splats and camera motion trajectory by integrating RGB loss to align the synthesized and captured blurry images, as well as event loss to ensure consistency between the synthesized and captured events along the trajectory.}
    \label{fig:method-scheme}
\end{figure*}
\section{Related Work}
\label{sec:related_work}

\subsection{Novel View Synthesis}
NeRF~\cite{mildenhall2020nerf} has garnered significant attention in the field of 3D vision due to its remarkable photo-realistic view synthesis capabilities.
At its core, NeRF employs a neural implicit representation optimized through a differentiable volume rendering technique.
Subsequent works have sought to enhance its rendering quality~\cite{barron2021mip,barron2022mip,jiang2022alignerf,wu2021diver,kaizhang2020nerfplusplus}, while other research directions focus on addressing its computational inefficiency by accelerating training and rendering~\cite{kangle2021dsnerf,lindell2021autoint,Reiser2021ICCV,chan2022efficient,yu2021plenoctrees,wang2022r2l,chen2023mobilenerf,fridovich2022plenoxels,muller2022instant,chen2022tensorf,sun2022dvgo}, achieving substantial improvements in speed.
More recently, 3DGS~\cite{kerbl20233d} has emerged as a refined variant of radiance field models, excelling in both detailed scene reconstruction and real-time rendering.
By replacing NeRF's computationally intensive ray marching~\cite{maxraymarching} with an efficient and deterministic rasterization approach, 3DGS maintains visual fidelity while enabling real-time rendering.
Leveraging the 3DGS framework, our approach aims to restore fine-grained details of latent sharp scenes from a blurry observation complemented by a corresponding event stream as presented in \figref{fig:method-overall}, while enabling real-time rendering of the reconstructed scenes.
\subsection{Single Image Deblurring}
A blurred image can be mathematically expressed as the convolution of a sharp image with a blur kernel. Classical methods~\cite{levin2009understanding, jia2007single, joshi2009image, shan2008high, xu2010two} typically address the deblurring problem by jointly optimizing the blur kernel and the underlying latent sharp image.
With the advent of deep learning, numerous end-to-end learning-based deblurring methods have been introduced~\cite{tao2018scalerecurrent, kupyn2019deblurganv2a, zamir2022restormera, nah2017deep, kupyn2018deblurgan, sun2015learning, jin2018learning}, consistently delivering superior qualitative and quantitative results.
However, these methods often rely on large paired datasets of blurry and sharp images for training, which limits their generalization performance when applied to domain-shifted images.
In contrast, our method, being a test-time optimization approach based on 3DGS, is inherently free from such generalization limitations. 
\subsection{Event Enhanced Image Deblurring}
Event cameras excel at capturing high temporal dynamic information \cite{gallego2022eventbased}, and many existing approaches leverage this capability to improve image deblurring results \cite{pan2019bringing, sun2022eventbaseda, wang2020event, sun2023event, xu2021motion, jiang2020learning, zhang2023generalizing}.
EDI \cite{pan2019bringing} presents a straightforward yet powerful approach, capable of generating sharp video sequences from various types of blur by solving a single-variable, non-convex optimization problem. Unlike EDI \cite{pan2019bringing}, methods such as \cite{sun2022eventbaseda, wang2020event, sun2023event} employ end-to-end neural networks to integrate event data for enhanced image deblurring and frame interpolation. A key innovation of our method lies in its ability to extract latent camera motion trajectory from the event stream, offering richer geometric constraints.
\section{Methodology}
\label{sec:method}
BeSplat aims to recover a sharp 3D scene representation by jointly learning the camera motion trajectories and Gaussians parameters, given a single motion-blurred image with corresponding event stream. This is achieved by minimizing the photometric error between the input blurred image and the synthesized blurred image, generated based on the physical motion blur image formation model, along with temporal event loss by incorporating event generation model as presented in \figref{fig:method-scheme}. We will deliver each content in the following sections.

\subsection{Preliminary: 3D Gaussian Splatting}
\label{3d-gs}
In 3D-GS \cite{kerbl20233d}, a scene is represented as a collection of 3D Gaussians, where each Gaussian \(\bG\) is characterized by its mean position \(\bmu \in \mathbb{R}^3\), 3D covariance matrix \(\mathbf{\Sigma} \in \mathbb{R}^{3 \times 3}\), opacity \(\bo \in \mathbb{R}\), and color \(\bc \in \mathbb{R}^3\). The scaled Gaussian distribution is given by:
\begin{equation}
    \bG(\bx) = e^{-\frac{1}{2}(\bx-\bmu)^{\top}\mathbf{\Sigma}^{-1}(\bx-\bmu)}. \label{eq:gauss}
\end{equation}

The covariance \(\mathbf{\Sigma}\) is parameterized using a scale \(\bS \in \mathbb{R}^3\) and rotation matrix \(\bR \in \mathbb{R}^{3 \times 3}\), represented as a quaternion \(\bq \in \mathbb{R}^4\), ensuring it remains positive semi-definite:
\begin{equation}
    \mathbf{\Sigma} = \mathbf{RSS}^{T}\mathbf{R}^{T}. \label{eq:3dcovariance}
\end{equation}

For rendering, 3D Gaussians are projected into 2D space from a camera pose \(\mathbf{T}_c \in \mathbb{R}^{3 \times 3}\) using:
\begin{align}
    \mathbf{\Sigma^{\prime}} &= \mathbf{JT}_c\mathbf{\Sigma T}_c^T\mathbf{J}^{T}, \quad 
    \mathbf{m}' = \frac{\mathbf{KTm}}{D}, \label{eq:covariance2d}
\end{align}
where \(\mathbf{\Sigma^{\prime}} \in \mathbb{R}^{2 \times 2}\) is the 2D covariance, \(D\) is the z-axis coordinate of \(\mathbf{m}\), and \(\mathbf{J} \in \mathbb{R}^{2 \times 3}\) is the Jacobian matrix.

Pixel colors are rendered by rasterizing the sorted 2D Gaussians based on depth:
\begin{equation}
    \bC = \sum_{i=1}^{N} \bc_i \alpha_i \prod_{j=1}^{i-1}(1-\alpha_j), \label{eq:render}
\end{equation}
where \(\bc_i\) is the color of the \(i\)-th Gaussian, and \(\alpha_i\) is its alpha value:
\begin{align}
    \alpha_i = \bo_i \cdot \exp(-\sigma_i), \quad \sigma_i = \frac{1}{2} {\rm \Delta}_i^T \mathbf{\Sigma^{\prime}}^{-1} {\rm \Delta}_i. \label{eq_alpha}
\end{align}
Here, \({\rm \Delta}_i \in \mathbb{R}^2\) is the offset between the pixel center and the 2D Gaussian center. The rendered pixel color \(\bC\) is differentiable with respect to all Gaussian parameters \(\bG\) and the camera poses \(\mathbf{T}_c\).

\subsection{Camera Motion Trajectory Modeling}
\label{subsec:camera_trajectroy}
Following the approach of Deblur-GS \cite{chen2024deblur}, we model the camera motion trajectory using a differentiable Bézier spline of order 7 in SE(3) space. This spline is parameterized by a set of learnable control knots, denoted as $\boldsymbol{T}_{c_i}^w \in \mathbb{SE}(3)$, where each $\boldsymbol{T}_{c_i}^w$ represents the transformation matrix from the camera's coordinate frame to the world frame for the \(i^{th}\) control knot.

For simplicity, we refer to $\boldsymbol{T}_{c_i}^w$ as $\boldsymbol{T}_i$ in subsequent derivations. These control knots are sampled at a uniform time interval, $\Delta t$, starting from \(t_0\). A smaller \(\Delta t\) provides a smoother representation of the motion trajectory but increases the number of control knots to optimize. Since the spline's value at a given timestamp is determined by four consecutive control knots, the starting index of these knots for a specific time \(t\) is computed as:
\begin{equation}
	k = \lfloor \frac{t - t_0}{\Delta t} \rfloor,
\end{equation}
where \(\lfloor * \rfloor\) is the floor operator. The four relevant control knots are \(\boldsymbol{T}_k\), \(\boldsymbol{T}_{k+1}\), \(\boldsymbol{T}_{k+2}\), and \(\boldsymbol{T}_{k+3}\).

Next, we define \(u = \frac{t - t_0}{\Delta t} - k\), where \(u \in [0, 1)\), to transform \(t\) into a normalized time representation. Using this representation, we calculate the cumulative basis matrix \(\mathcal{B}(u)\) based on the De Boor-Cox formula \cite{qin1998general}:
\begin{equation}
	\mathcal{B}(u) = \mathcal{M} \begin{bmatrix}
		1 \\ u \\ u^2 \\ u^3
	\end{bmatrix}, \quad
	\mathcal{M} = \frac{1}{6} \begin{bmatrix}
		6 & 0 &  0 &  0 \\
		5 & 3 & -3 &  1 \\
		1 & 3 &  3 & -2 \\
		0 & 0 &  0 &  1
	\end{bmatrix}.
\end{equation}
Finally, the camera pose at time \(t\) is derived as:
\begin{equation}\label{eq_rgb}
	\boldsymbol{T}(u) = \boldsymbol{T}_k \cdot \prod_{j=0}^2 \mathrm{exp}(\mathcal{B}(u)_{j+1} \cdot \bOmega_{k+j}),
\end{equation}
where \(\mathcal{B}(u)_{j+1}\) represents the \((j+1)^{th}\) element of the vector \(\mathcal{B}(u)\), and \(\bOmega_{k+j} = \mathrm{log}(\boldsymbol{T}_{k+j}^{-1} \cdot \boldsymbol{T}_{k+j+1})\).

This formulation provides a differentiable and flexible representation of the camera trajectory, enabling optimization for accurate motion modeling.

\section{Differentiable Pose Estimation}
We derive the motivation from Deblur-GS \cite{chen2024deblur} for the differentiable pose estimation framework. The gradient of loss \(L\) with respect to the camera pose parameters \(T\), $\frac{\partial L}{\partial T}$ can be expressed as:
\[
\frac{\partial L}{\partial \hat{C}} 
\left[
\sum_{i \in N} \frac{\partial \hat{C}}{\partial c_i} \frac{\partial c_i}{\partial T} + 
\sum_{i \in N} \frac{\partial \hat{C}}{\partial \alpha_i} 
\left( 
\frac{\partial \alpha_i}{\partial \Sigma'_i} \frac{\partial \Sigma'_i}{\partial T} + 
\frac{\partial \alpha_i}{\partial m'_i} \frac{\partial m'_i}{\partial T}
\right) 
\right]
\]

The gradient with respect to the camera pose is simplified by retaining only the Position Term, as the Color and Covariance Terms are negligible. This leads to the final expression:

\[
\frac{\partial L}{\partial T} = 
\sum_{i \in N} \frac{\partial L}{\partial \hat{C}} 
\frac{\partial \hat{C}}{\partial \alpha_i} 
\frac{\partial \alpha_i}{\partial m'_i} 
\frac{\partial (K T m_i)}{\partial T D_i}.
\]

The explicit representation of 3D Gaussian points allows the camera pose transformations to be handled as transformations of the Gaussian point positions:
\[
K(T'T)m_i = K T (T'm_i)
\]
where \(T'\) represents an external transformation applied to the camera pose \(T\). This is equivalent to directly applying \(T'\) to the positions of the Gaussian points.

By optimizing a global transformation for all Gaussian points for each camera, the pre-transformed Gaussian point positions \(\hat{m}_i = T m_i\) are passed directly to the differentiable renderer without further modification. The renderer then computes the gradient of \(L\) with respect to \(\hat{m}\), while PyTorch's automatic differentiation handles the computation of the gradient of \(\hat{m}\) with respect to \(T\).

The desired gradient is ultimately calculated as:
\[
\frac{\partial L}{\partial T} = 
\frac{\partial L}{\partial \hat{m}}
\frac{\partial \hat{m}}{\partial T}
\]

\subsection{Blurry Image Formation Model}
A motion-blurred image $\bB(\bx) \in \nR^{\mathrm{W} \times \mathrm{H} \times 3}$ is generated by accumulating photons over the exposure duration, and it can be mathematically expressed as:
\begin{equation} \label{eq_blurry}
	\bB(\bx)  \approx \frac{1}{n} \sum_{i=0}^{n-1} \bI_\mathrm{i}(\bx), 
\end{equation}
where $\mathrm{W}$ and $\mathrm{H}$ denote the width and height of the image, respectively, $n$ represents the number of sampled images, and $\bx \in \nR^2$ specifies the pixel location. Here, $\bI_\mathrm{i}(\bx) \in \nR^{\mathrm{W} \times \mathrm{H} \times 3}$ corresponds to the $i^{th}$ virtual sharp image sampled during the exposure time. These virtual sharp images are rendered from the neural 3D scene representation, following the previously defined camera trajectory. Notably, $\bB(\bx)$ is differentiable with respect to the parameters of both the 3D Gaussians and the motion trajectory.

\subsection{Event Data Formation Model}
Event cameras asynchronously record brightness changes as events. An event $e_i = (\bx, t_i, p_i)$ is triggered when the brightness at pixel $\bx$ changes by more than a contrast threshold $C$ (\ie $|L(\bx, t_i + \delta t) - L(\bx, t_i)| \ge C$), where $p_i \in \{-1, +1\}$ represents the polarity, and $L(\bx, t_i) = \log(\bI(\bx, t_i))$ is the logarithmic brightness.

To relate the radiance field and event stream, events within a time interval $\Delta t$ are accumulated into an image:
\begin{equation}\label{eq_event_accumulate}
	\bE(\bx) = C \{e_i(\bx, t_i, p_i)\}_{t_k < t_i < t_k + \Delta t}.
\end{equation}
Since the contrast threshold $C$ varies, we normalize the accumulated events as:
\begin{equation}\label{eq_event_normalize}
	\bE_n(\bx) = \frac{\bE(\bx)}{\norm{\bE(\bx)}_2}.
\end{equation}

Using the interpolated start and end poses from a Bézier curve, two grayscale images, $\bI_{start}$ and $\bI_{end}$, are rendered via 3D Gaussian Splatting. The synthesized accumulated event image is:
\begin{equation}
	\hat{\bE}(\bx) = \log(\bI_{end}(\bx)) - \log(\bI_{start}(\bx)).
\end{equation}
This result, dependent on the cubic Bézier curve and 3D Gaussians, is differentiable and can be normalized as in \eqref{eq_event_normalize} for loss computation.

\subsection{Loss Functions}
We minimize a combined objective function, consisting of a photometric loss \(\cL_p\) and an event loss \(\cL_e\):  
\begin{equation}
    \cL_{total} = \alpha\cL_p + \beta\cL_e,
\end{equation}
where \(\alpha\) and \(\beta\) are weighting hyperparameters. The photometric loss \(\cL_p\) measures the reconstruction error for frame-based images, while the event loss \(\cL_e\) quantifies the discrepancy for accumulated events over a sampled time interval:  
\begin{align}
	&\cL_p = \norm{\bB(\bx) - \hat{\bB}(\bx)}^2, \\
	&\cL_e = \norm{\bE_n(\bx) - \hat{\bE}_n(\bx)}^2.
\end{align}
Here, \(\bB(\bx)\) and \(\hat{\bB}(\bx)\) are the captured and synthesized blurry images, while \(\bE_n(\bx)\) and \(\hat{\bE}_n(\bx)\) are the normalized and synthesized event images. This joint optimization aligns frame and event based modalities, improving reconstruction quality.

\section{Experiments}
\label{sec:experiments}
\subsection{Experimental Setup}
\PAR{Datasets:} 
We use synthetic dataset generated by BeNeRF\cite{li2025benerf} via Unreal Engine \cite{unreal} and Blender\cite{foundationblender}  for evaluations. In total, dataset contains three sequences (\ie livingroom, whiteroom and pinkcastle) via Unreal Engine and two sequences (\ie tanabata and outdoorpool) via Blender. The event streams are generated via ESIM \cite{rebecq2018esim} from high frame-rate video. We employed the real-world dataset introduced by E$^2$NeRF, captured using the DAVIS346 color event camera. This dataset comprise five challenging scenes (\ie letter, lego, camera, plant, and toys), each characterized by intricate textures and diverse motion patterns. The RGB frames were recorded with an exposure time of 100ms, leading to instances of complex camera trajectories and significant motion blur during the capture interval.

\PAR{Baseline methods and Evaluation metrics:}
To assess the effectiveness of our method in terms of image deblurring, we compare it with state-of-the-art deep learning-based single-image deblurring techniques, including DeblurGANv2 \cite{kupyn2019deblurganv2a}, MPRNet \cite{zamir2021multistage}, NAFNet \cite{chen2022simple}, Restormer \cite{zamir2022restormera}, event-enhanced single-image deblurring method EDI \cite{pan2019bringing} and BeNeRF\cite{li2025benerf}.Our primary contribution lies in achieving faster training times and enabling real-time rendering. To comprehensively evaluate our approach, we report memory usage and GPU hours alongside standard image reconstruction metrics, including Peak Signal-to-Noise Ratio (PSNR), Structural Similarity Index Measure (SSIM), and Learned Perceptual Image Patch Similarity (LPIPS).

\PAR{Implementation Details:}
BeSplat is implemented using the PyTorch framework. The Gaussian scene parameters and camera pose parameters are optimized using two separate Adam optimizers, as outlined in \cite{chen2024deblur}. The learning rate for the pose optimizer decays exponentially from \(1 \times 10^{-3}\) to \(1 \times 10^{-5}\). For the real dataset, we set \(\alpha=0.1\) and \(\beta=1.0\), while for the synthetic dataset, \(\alpha=1.0\) and \(\beta=2.0\). A 7th-order Bézier interpolation is employed with 19 sample points along the camera motion trajectory. For initializing Gaussian points and control points, we utilize COLMAP \cite{schonberger2016structure}. The model is trained for 30,000 iterations on a single NVIDIA RTX 3090 GPU. Initially, the control points of the camera trajectory overlap, but they are progressively optimized to distinct positions during training. To obtain the initial sparse point cloud and camera pose for both real and synthetic datasets, structure-from-motion (COLMAP) is applied to the corresponding sharp images, following the approach in \cite{chen2024deblur}.

\subsection{Qualitative Evaluations}
\begin{figure*}[th]
    \centering
    \includegraphics[width=0.83\linewidth]{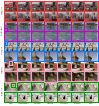}
    \caption{\textbf{Results on the Synthetic Dataset:} Our method consistently achieves high performance, closely rivaling BeNeRF in image reconstruction quality. Insets highlight specific regions of the images, demonstrating that our method achieves comparable visual fidelity to BeNeRF while offering significant reductions in training and rendering times, as well as lower GPU memory usage.}
    \label{fig_synthetic}
\end{figure*}

\begin{figure*}[th]
    \centering
    \includegraphics[width=0.85\linewidth]{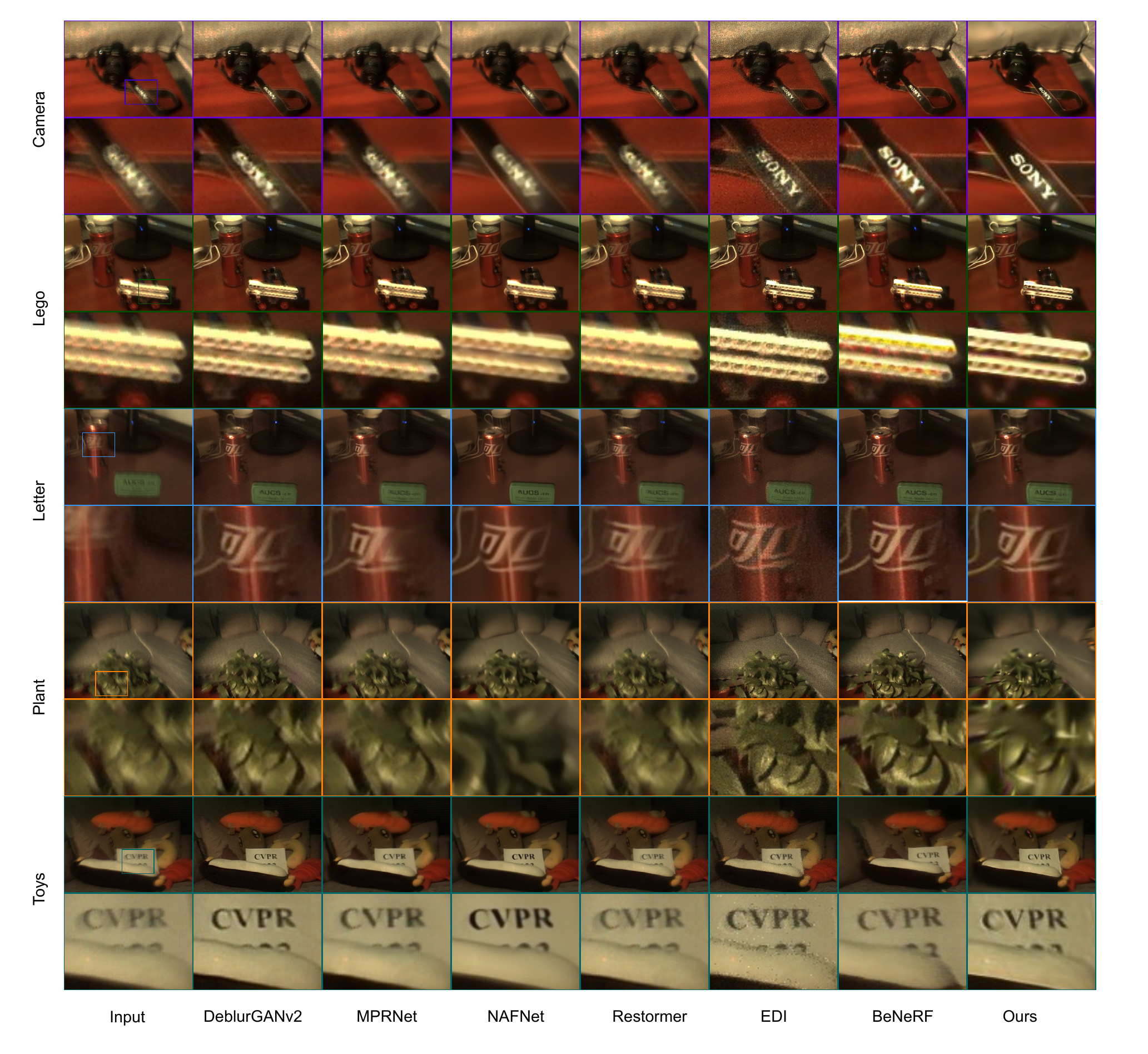}
    \caption{\textbf{Results on the Real Dataset:} Our method effectively handles varying levels of blur, consistently producing sharp, high-quality images. Insets illustrate that while our approach delivers results comparable to state-of-the-art methods on synthetic datasets, it performs noticeably better on real-world data by providing sharper and cleaner details, highlighting its robustness and reliability.}
    \label{fig_real}
\end{figure*}

The qualitative evaluation results, presented in \figref{fig_synthetic} and \figref{fig_real}, demonstrate the performance of our method on both synthetic and real datasets.Although BeNeRF, as a NeRF-based approach, achieves slightly better performance due to its detailed neural representation, our method, which leverages Gaussian Splatting, delivers high-quality results with the added benefits of faster training times, real-time rendering capabilities, and reduced GPU memory usage. Specifically, \figref{fig_synthetic} showcases how our method consistently preserves reconstruction quality across diverse scenarios. Additionally, \figref{fig_real} highlights the robustness of our approach on real noisy datasets, achieving competitive results and maintaining high visual fidelity despite challenging conditions.

\subsection{Quantitative Evaluations}
Given our focus on accelerated training times, reduced GPU memory usage, and real-time rendering performance while maintaining high-quality reconstructions, we present metrics that comprehensively evaluate these aspects alongside standard image reconstruction quality measures such as PSNR, SSIM, and LPIPS. As illustrated in \tabnref{table_memory_cost}, our method achieves a remarkable reduction in GPU memory consumption, highlighting its suitability for resource-constrained environments. Additionally, \tabnref{table_training_time} showcases the significantly shorter training times of our approach compared to BeNeRF \cite{li2025benerf}, underscoring its computational efficiency. Furthermore, as detailed in \tabnref{table_psnr_ssim_lpips}, our method demonstrates competitive performance in image reconstruction quality metrics, performing on par with state-of-the-art methods like BeNeRF \cite{li2025benerf}.

\begin{table}[t]
    \centering
   \caption{\textbf{Quantitative Comparisons:} The table evaluates image deblurring methods using PSNR, SSIM, and LPIPS metrics. Our method delivers reliable performance with a focus on efficiency and resource optimization.}
    \setlength\tabcolsep{6.5pt}
    \resizebox{0.9\linewidth}{!}{
        \begin{tabular}{c|ccc|ccc}
            & \multicolumn{3}{c|}{Livingroom} & \multicolumn{3}{c}{Tanabata}\\
            & PSNR$\uparrow$ & SSIM$\uparrow$ & LPIPS$\downarrow$ & PSNR$\uparrow$ & SSIM$\uparrow$ & LPIPS$\downarrow$ \\
            \specialrule{0.05em}{1pt}{1pt}
            DeblurGANv2 \cite{kupyn2019deblurganv2a} & 29.26 & .8121 & .2087 & 20.09 & .4964 & .3934 \\
            MPRNet \cite{zamir2021multistage} & 28.57 & .7937 & .2621 & 18.20 & .4258 & .4173 \\
            NAFNet \cite{chen2022simple} & 29.92 & .8306 & .2268 & 18.96 & .4665 & .3908 \\
            Restormer \cite{zamir2022restormera} & 29.48 & .8262 & .2391 & 18.82 & .4596 & .4248 \\
            EDI\cite{pan2019bringing} & 32.61 & .8871 & \cellcolor{yellow!25}.0904 & 24.87 & .7564 & .1039 \\
            BeNeRF \cite{li2025benerf} & \cellcolor{red!25}37.11 & \cellcolor{red!25}.9370 & \cellcolor{red!25}.0632 & \cellcolor{red!25}32.14 & \cellcolor{red!25}.9015 & \cellcolor{red!25}.0515 \\
            \textbf{Ours} & \cellcolor{yellow!25}35.14 & \cellcolor{yellow!25}.9111&  .1189 & \cellcolor{yellow!25}29.15 & \cellcolor{yellow!25}.8626 & \cellcolor{yellow!25}.1015 \\
            \specialrule{0.05em}{1pt}{1pt}
        \end{tabular}
    }
    \label{table_psnr_ssim_lpips}
\end{table}
\begin{table}[t]
    \centering
    \caption{{\textbf{Memory Usage Comparison with BeNeRF on Real Dataset (GB):}} Our method significantly reduces memory consumption compared to BeNeRF, showcasing its efficiency.}
    \resizebox{\columnwidth}{!}{
        \begin{tabular}{c|cccccc}
            \hline
            & Camera$\downarrow$ & Lego$\downarrow$ & Letter$\downarrow$ & Plant$\downarrow$ & Toys$\downarrow$ & \textit{Average}$\downarrow$ \\
            \specialrule{0.05em}{1pt}{1pt}
            
            BeNeRF\cite{li2025benerf} & \cellcolor{yellow!25}$8.65$ & 
            \cellcolor{yellow!25}$8.65$ & \cellcolor{yellow!25}$8.65$ & \cellcolor{yellow!25}$8.65$ & \cellcolor{yellow!25}$8.65$ & \cellcolor{yellow!25}$8.65$ \\
           
            \textbf{Ours} & \cellcolor{red!25}$1.45$ & 
             \cellcolor{red!25}$1.45$ & \cellcolor{red!25}$1.45$ & \cellcolor{red!25}$1.45$ & \cellcolor{red!25}$1.45$ & \cellcolor{red!25}$1.45$ \\
            \specialrule{0.05em}{1pt}{1pt}
        \end{tabular}
    }
    \label{table_memory_cost}
\end{table}
\begin{table}[t]
    \centering
    \caption{{\textbf{Training Time Comparison with BeNeRF on Real Dataset (hh:mm):}} Our method demonstrates a substantial reduction in training time compared to BeNeRF, highlighting its efficiency.}
    \resizebox{\columnwidth}{!}{
        \begin{tabular}{c|cccccc}
            \hline
            & Camera$\downarrow$ & Lego$\downarrow$ & Letter$\downarrow$ & Plant$\downarrow$ & Toys$\downarrow$ & \textit{Average}$\downarrow$ \\
            \specialrule{0.05em}{1pt}{1pt}
            
            BeNeRF\cite{li2025benerf} & \cellcolor{yellow!25}$06.20$ & \cellcolor{yellow!25}$06.25$ & \cellcolor{yellow!25}$06.25$ & \cellcolor{yellow!25}$06.30$ & \cellcolor{yellow!25}$06.25$ & \cellcolor{yellow!25}$06.25$ \\
            \textbf{Ours} & \cellcolor{red!25}$01.30$ & \cellcolor{red!25}$01.25$ & \cellcolor{red!25}$01.30$ & \cellcolor{red!25}$01.35$ & \cellcolor{red!25}$01.30$ & \cellcolor{red!25}$01.30$ \\
            
            \specialrule{0.05em}{1pt}{1pt}
        \end{tabular}
    }
    \label{table_training_time}
\end{table}

\begin{figure}[th]
    \centering
    \includegraphics[width=0.81\columnwidth]{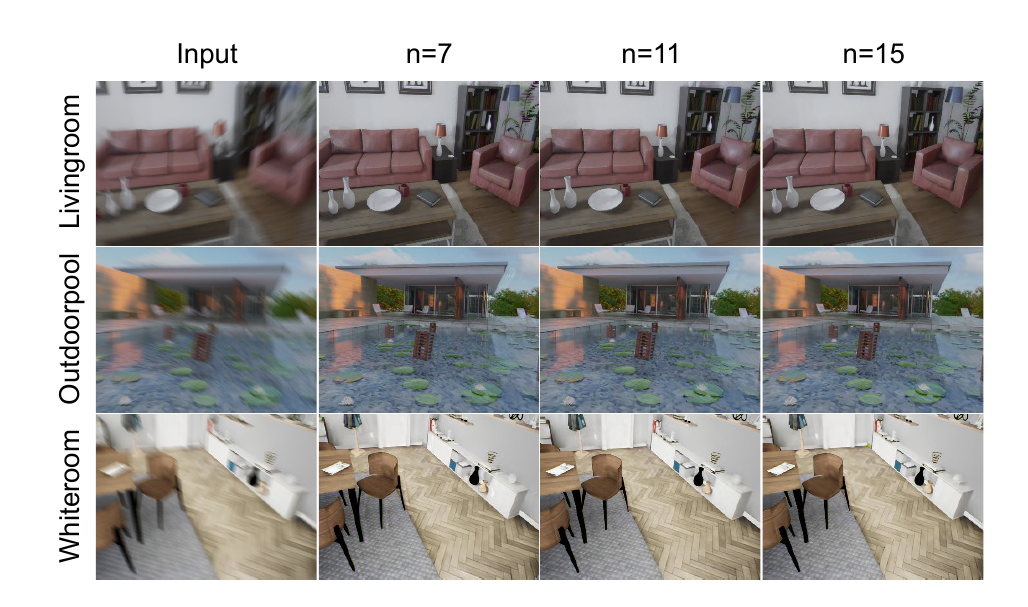}
 \caption{\textbf{Ablation Studies on Virtual Camera Count:} Our method demonstrates robust deblurring performance, maintaining high-quality results even when using a limited number of virtual cameras along the camera motion trajectory.}
    \label{fig_ablation}
\end{figure}

\subsection{Ablation Study}
We conducted ablation studies to assess the impact of trajectory representations and the number of virtual cameras used for optimization along the camera motion trajectory.
As shown in \figref{fig_ablation}, our method effectively reconstructs sharp images even with a limited number of virtual cameras, demonstrating its robustness in sparse setups. Additionally, we evaluated different motion trajectory models, including linear interpolation and cubic B-splines.
The results, summarized in \tabnref{table_livingroom_tanabata}, indicate that the Bézier curve model achieves slightly better reconstruction quality compared to cubic B-splines, with higher PSNR and SSIM values and slightly lower LPIPS scores across both the Livingroom and Tanabata datasets. While exhibiting comparable reconstruction quality to Bézier curves, cubic B-splines necessitate significantly longer training times, typically requiring approximately three times the computational effort for optimization. This makes the Bézier model a more efficient and practical choice for trajectory representation, delivering 
reliable performance with reduced computational overhead.

\begin{table}[t]
    \centering
    \caption{\textbf{Ablation Studies on Trajectory Representations:} Bézier curves exhibit superior accuracy to B-splines \& linear interpolation with significantly faster training, making them ideal for efficient trajectory representation.}
    \setlength\tabcolsep{6.5pt}
    \resizebox{0.9\linewidth}{!}{
        \begin{tabular}{c|ccc|ccc}
            & \multicolumn{3}{c|}{Livingroom} & \multicolumn{3}{c}{Tanabata}\\
            & PSNR$\uparrow$ & SSIM$\uparrow$ & LPIPS$\downarrow$ & PSNR$\uparrow$ & SSIM$\uparrow$ & LPIPS$\downarrow$ \\
            \specialrule{0.05em}{1pt}{1pt}
            linear interpolation  & 32.08 & .8758 & .1715 & 25.25 & .7747 & .1732 \\
            cubic B-Spline & \cellcolor{yellow!25}34.99 & \cellcolor{yellow!25}.9096 & \cellcolor{yellow!25}.1204 & \cellcolor{yellow!25}29.07 & \cellcolor{yellow!25}.8599 & \cellcolor{yellow!25}.1027 \\
            \textbf{Bézier(Ours)}  & \cellcolor{red!25}35.14 & \cellcolor{red!25}.9111 & \cellcolor{red!25}.1189 & \cellcolor{red!25}29.15 & \cellcolor{red!25}.8627 & \cellcolor{red!25}.1015 \\
            \specialrule{0.05em}{1pt}{1pt}
        \end{tabular}
    }
    \label{table_livingroom_tanabata}
\end{table}

\section{Conclusion}
\label{sec:conclusion}
We introduce a novel method based on the Gaussian Splatting framework that simultaneously recovers the underlying 3D scene representation and estimates the precise camera motion trajectory from a single blurry image and its associated event stream. Through extensive evaluations on synthetic and real-world datasets, our approach demonstrates a compelling balance of efficiency and quality, achieving accelerated training times, reduced GPU memory usage, and real-time rendering capabilities, all while delivering high-quality reconstructions that validate its practical robustness and versatility. By leveraging the complementary event stream data, our method effectively captures motion details and enhances reconstruction fidelity. Future work could explore extending the framework to handle dynamic scenes.

\textbf{Acknowledgement:} 
 We would like to acknowledge partial support from the Institute of Eminence (IoE) Research Center on VR and Haptics, together with the IITM Pravartak Technologies Foundation. KM would like to acknowledge support from Qualcomm Faculty Award 2024.



\end{document}